# Semantic Segmentation by Improved Generative Adversarial Networks

ZengShun Zhao[a,*], Yulong Wang[a], Ke Liu[a], Haoran Yang[a], Qian Sun[a],

*Heng Qiao* [b]

[a] *College of Electronic and Information Engineering,*

*Shandong University of Science and Technology, Qingdao, 266590, P.R.China*

*Correspondence Author: zhaozs@sdust.edu.cn;zhaozengshun@163.com
*Co-Correspondence Author:17854252591@163.com， sunqian940411@163.com

[b] *Department of Electrical& Computer Engineering, University of Florida,*

*Gainesville, FL 32611,USA*

**Abstract**

While most existing segmentation methods usually combined the powerful feature extraction capabilities of CNNs with Conditional Random Fields (CRFs) post-processing, the result always limited by the fault of CRFs . Due to the notoriously slow calculation speeds and poor efficiency of CRFs, in recent years, CRFs post-processing has been gradually eliminated. In this paper, an improved Generative Adversarial Networks (GANs) for image semantic segmentation task (semantic segmentation by GANs, Seg-GAN) is proposed to facilitate further segmentation research. In addition, we introduce Convolutional CRFs (ConvCRFs) as an effective improvement solution for the image semantic segmentation task. Towards the goal of differentiating the segmentation results from the ground truth distribution and improving the details of the output images, the proposed discriminator network is specially designed in a full convolutional manner combined with cascaded ConvCRFs. Besides, the adversarial loss aggressively encourages the output image to be close to the distribution of the ground truth. Our method not only learns an end-to-end mapping from input image to corresponding output image, but also learns a loss function to train this mapping. The experiments show that our method achieves better performance than state-of-the-art methods.

---

**1 Introduction**

Image semantic segmentation has become one of the noteworthy and active research eares in the filed of computer vision and computer graphics, and attracted increasing attention in many application of daily life, such as autopilot [1], medical image analysis [2], geographic information system and smart dressing system. The core task of semantic segmentation is to mark a semantic label, e.g., table, bus, plane, person ,or horse, to each pixel in images. In recent years, while many recent approaches[] have been proposed to tackle this task, and many popular datasets have been built for the research of semantic segmentation, image semantic segmentation is still facing challenges.

Despite the accuracy of semantic segmentation based on deep learning methods are far superior to the traditional methods in extracting local features and performing good predictions utilizing small field of view, deep neural networks lack the capability to utilize global context information and slow calculations during training. In addition, the long training times of the current generation of CRFs in post-processing also make more in-depth research and experiments with such structured models impractical.

For typically CNN-based segmentation networks, the corresponding ground truths and their original images in the training dataset are used to train the segmentation network, and the constant training of the network is guided by directly comparing the differences between the segmented results and the ground truths.

In this paper, we applied a deep learning based approach for image semantic segmentation. More specifically, we proposed Seg-GANs for this task, which was inspired by the GANs and ConvCRFs. Similar to the standard GANs, Seg-GAN also consists of two feed-forward convolutional neural networks (CNNs), the segmentation network S and the discriminator network D. The segmentation network plays a similar role of the generative network in the original GANs, and the goal of the segmentation network S is to generate segmentation results from the input images and assign the labels for every pixel. The discriminative network D aims to discover the discrepancy between the segmented results and the corresponding ground-truth image. Besides, the discriminative network adopts four cascaded ConvCRFs [4] layers to facilitate the fully connected network to have modest accuracy and inference speed improvement. In addition, a focal loss function is introduced to calculate the confidence map generated by the discriminative network. The loss function,including two-class cross-entropy error loss and adversarial loss,guides the constant training of the discriminative network and improves the accuracy. In this paper, we demonstrate that the proposed networks are effective in the segmentation task. Experiments show that our method outperforms current state-of-the-art methods both perceptually and quantitatively.

Our proposed method differs from the existing traditional [5, 6] or other deep learning based approaches [7-12]. The traditional approaches need to extract the features of the images manually. The deep learning based approaches are usually based on CNNs.

There are three main innovations in our work:

(1) We proposed Seg-GANs, an end-to-end generative adversarial network, for image semantic segmentation based on residual networks framework. In our algorithm, the generative network counts on a remarkable basement segmentation network to generate the class prediction image. Normally, CRFs based segmentation network can considerable improve the accuracy[9-11], but with the risk of decreasing the computation efficiency and hindering the construction of an end-to-end framework. We abandoned the CRFs post-processing module in the segmentation network and put CRFs into the end-to-end training to improve the accuracy.

(2) Based on the structure of GAN, Seg-GANs combine two-class cross-entropy error loss to calculate loss function. The discriminative network we designed consists of a fully connected networks with four cascade ConvCRFs layer and given up the multi-scale fusion method, and the output of the discriminative network substitute by the confidence map according to our networks rather than a simple loss value. And then we can control the discriminator to input image of any size. Each value in the confidence map is sampled from a different region of the input and represents the confidence values of all the segmentation targets in the image.

(3) ConvCRFs have indisputable success in the speeding up inference and training as described in research [4]. We first introduce ConvCRFs to replace the CRFs to modify this problem, in our algorithm, by fusing ConvCRFs layer with the discriminative network to improve the efficiency of calculation. We further demonstrate that Generative Adversarial Networks are useful in the image semantic segmentation task, and can achieve better accuracy than the other deep learning based methods. Our method directly learns an end-to-end mapping which can effectively estimate the reasonable results from input images and make the calculation more effectively.

**2 Related Works**

In the early studies, various constructive methods have been proposed for image semantic segmentation, such as threshold-based, region-based [5], edge-based [6] and cluster-based techniques are proposed for segmenting the image. While these traditional methods simply separate objects from the background, this process has to manually design a large number of features. The quality of the several features directly determines the quality of the segmentation results, which is time-consuming and labor-intensive and not practical enough.

In recent years, with the increasing research on convolutional neural networks and deep learning framework in image semantic segemetation [13,14], more and more studies related to deep convolutional neural networks has been carried out to increasingly improve the semantic segmentation methods, which are expected to ameliorate the accuracy of recognition and prediction. Long et al. [7] were the first to applie deep convolutional neural networks to the task of image semantic segmentation, which utilized the fully connected layer to replace the convolutional layer in FCN module. This fully convolutional neural network (FCN), as one of the most popular prototypes of the encoder-decoder framework, is adopted for pixel-level image classification. By up-sampling with transposed convolution, full-size segmented image could be restored with classified pixels. With the improvement in GPU performance and optimization algorithms, researchers started to train larger and deeper neural networks.

Noe et al. proposed DeconvNet [15] with a more extensive decoder than the original FCN. The mentioned decoder is symmetric with respect to the number and feature sizes of the encoder. Aside from the deconvolution, the DeconvNet decoder network also uses unpooling layers as a part of improvement. Since the DeconvNet uses two fully-connected layers in its encoder, it is relatively large in memory consuming compared with the original FCN.

Motivated to reduce the number of parameters and the amount of memory required by segmentation networks, Badrinarayanan et al. propose the SegNet [1], which encoder is topologically identical to the 13 convolutional layers of VGG-16 [16], but in contrast to the original FCN and DeconvNet, the decoder contains only up-sampling (unpooling) operations and convolution, therefore eliminating deconvolution altogether. Architectures that store and use feature maps from an encoder during classification is outperformed SegNet but require more memory during inference.

The second method that is widely used in semantic segmentation is the dilated convolutional structure [17]. The DeepLab method has developed about four versions. The main contributions of the former two versions DeepLab-v1 [9] and DeepLab-v2 [10] are the combination of convolutional neural networks and fully connected CRFs and the model innovatively applies the dilated convolution algorithm to the convolutional neural network models. The biggest difference between DeepLab-v3 [11] and DeepLab-v3+ [12] with the previous two versions is that the CRFs post-processing module in DeepLab-v3 and DeepLab-v3+ is abandoned and substituted with the changed atrous spatial pyramid pooling (ASPP) [10] structure, but the cascading network structure inevitable puts tremendous pressure on GPU memory.

Recently, there are a large body of successful extended applications based on generative adversarial networks (GANs) (e.g., SRGAN [18], DCGAN [19], Pix2Pix [20]) since Goodfellow first officially proposed GANs in 2014. GANs perform an adversarial process alternating between identifying and faking, and the generative adversarial loss is formulated to evaluate the discrepancy between the generated distribution and the real data distribution. A lot of researches reveal that generative adversarial loss is beneficial for improving the performance of the networks. Inspired by the success of generative adversarial networks (GANs) on image-to-image translation [20], we designed an efficient GAN network for image semantic segmentation. The work closest in scope to ours is the one proposed by Luc et al. [8], where the adversarial network is used to aid the training for semantic segmentation. However, it does not show substantial improvement over the baseline.

To further introduce global information into CNNs, Deeplab uses the fully connected CRFs (FullCRFs) as an independent post-processing step. FullCRFs [21] utilize two Gaussian kernels with hand crafted features as illustrated in the original publication [21], Krähenbühl and Koltun optimized the remaining parameters with a combination of grid-search and expectation maximization. In the next work [22] they novelty used gradient decent that for the message passing the identity $(k_G * Q)^{'} = k_G * Q^{'}$ is valid. However, for the reason of using back propagation without computing gradients with respect to the Gaussian kernel $k_G$, the features of the Gaussian kernel therefore cannot be learned. The subsequently proposed CRFasRNN [23] adopts the same ideas to implement joint CRFs and CNN training and also requires hand-crafted Gaussian features like [22].

But the long training times of the current generation of CRFs make more in-depth research and experiments with such structured models impractical. In order to circumvent the issue of notoriously slow training and long inference times of CRFs, Teichmann M T T. et al. [4] developed Convolutional CRFs (ConvCRFs), a novel CRFs design, which adding the strong and valid assumption of conditional independence so as to remove the permutohedral lattice [24] approximation. Based on the validation experiments of this approach, this approach increases training and inference speed by two orders of magnitude. Besides, the ConvCRFs implementation utilizing a learnable compatibility transformation as well as learnable Gaussian features performs best and reformulating a large proportion of the inference as convolutions thus can be implemented highly efficiently on GPUs.

**3 Generative Adversarial Networks for image semantic segmentation**

In this section, we will introduce the proposed structure of Generative Adversarial Networks for image semantic segmentation. The main framework of proposed algorithm is shown in Fig. 1. Compared to original GAN model, the architecture of the proposed Seg-GAN is based on two separate deep convolutional neural networks, namely the segmentation network S and discriminator network D, whose combined efforts aim at obtaining a reasonable result for a given input image.

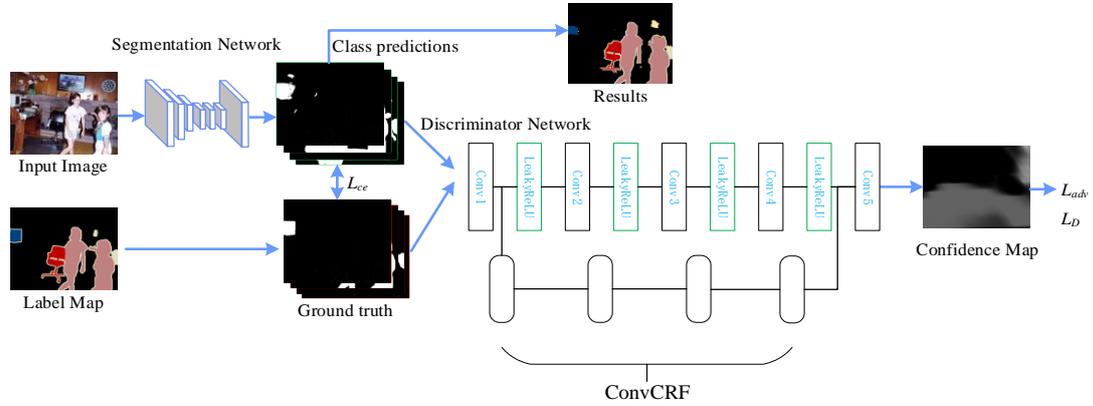

Figure 1: Architecture of the proposed Seg-GAN.

**3.1 Segmentation network**

The segmentation network S is designed for generating a reasonable result by segmenting the given input image. The structure of generative network is inspired by the configuration of DeepLab-v2 [10] framework with ResNet-101 [25] model pre-trained on MS COCO dataset which is our segmentation baseline network and without CRFs post-processing. For the GPU memory consuming consideration, we abandon to adopt the multi-scale fusion proposed in Chen et al [10]. Following the recent work on semantic segmentation, we drop the last classification layer and revise the stride of the last two convolution layers from 2 to 1, making the resolution of the output feature maps effectively $1/8$ times the input image size. Towards the goal of enlarging the receptive fields, we adopt the dilated convolution in conv4 and conv5 layers with a stride of 2 and 4, respectively. After the last layer, we use the Atrous Spatial Pyramid Pooling (ASPP) proposed in Chen et al. [10] as the final classifier.

Correspondingly, we apply an up-sampling layer along with the softmax output to adapt the size of the input image. The architecture of the segmentation network S is demonstrated in the Fig. 2.

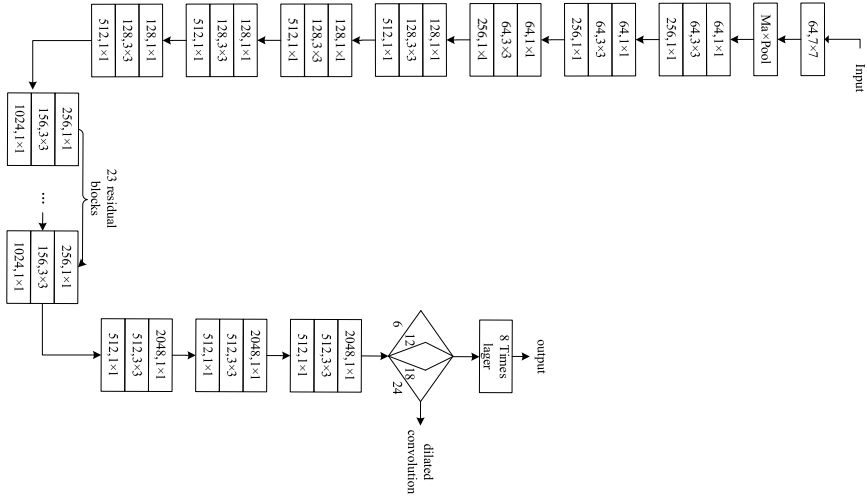

Figure 2: The architecture of the segmentation network S.

## 3.2 Discriminative network

The discriminative network D is proposed to compute the discrepancy between the data distribution of the ground-truth labels and the predicted label images generated by generative network. The proposed discriminator network is specially designed in a full convolutional manner combined with cascaded ConvCRFs to differentiate the segmentation results from the ground truth distribution and improve the details of the output images. It consists of 5 convolution layers with kernel $3\times3$ with channel numbers {64, 128, 256, 512, 1} and stride of 2. Each convolution layer is followed by a Leaky-ReLU parameterized by 0.2 except the last layer. We first use 4 ConvCRF modules cascaded with the full convolutional networks.

## 3.3 Generative adversarial loss

The GAN-based models have been widely used in learning generative model due to their indisputable success in image generation. The GANs was proposed to solve the disadvantages of other generative models. Instead of maximizing the possibility, GANs introduce the theory of adversarial learning between the generator and the discriminator. This adversarial process gives GANs obvious advantages over the other generative models. Moreover, GANs can sample the generated data in a simple way unlike other models in which the sampling is notoriously slow and inaccurate. For these advantages, GAN gained our attention, and this is the original intention for us to use the framework of GAN. We therefore adapt the GANs learning strategy to tackle the problem of image semantic segmentation. More specifically, the proposed Seg-GAN consists of two feed-forward convolutional neural networks (CNNs): the segmentation network S and the discriminator network D. The reason why we use CNN is that it can greatly stabilize GAN training. Seg-GAN suggests an architecture guideline in which the segmentation network is composed of a CNN, and the discriminator is composed of a full convolutional manner combined with cascaded ConvCRFs [4]. Batch normalization, ReLU and LeakyReLU activation functions are utilized for the segmentation network and the discriminator to help stabilize the GAN training.

The purpose of the segmentation network S is to generate labeled segmentation results $S(x)$ from input image $x$. Meanwhile, each input image $x$ has a corresponding ground-truth image $y$. $S(x)$ is encouraged to have the same data distribution with the ground-truth image $y$. The goal of the discriminator network D is to discover the discrepancy between the data distribution of segmentation results and the corresponding ground-truth image. S and D compete with each other to achieve their respective purposes, thus generate the term adversarial. To train the discriminator network, we minimize the cross-entropy loss $L_D$ with respect to two classes. The loss can be expressed as:

$$L_D = -\sum_{h,w}(1-y_n)\log\left(1-D\left(S(X_n)\right)^{(h,w)}\right) + y_n\log\left(D(Y_n)^{(h,w)}\right) \qquad (1)$$

In Eq. 1, $X_n$ is the input image with size of $h \times w \times 3$. We denote the segmentation network as $S(\square)$ which has a corresponding output $S(X_n)$. For our fully convolutional discriminator, we denote it as $D(\square)$. $Y_n$ is the corresponding ground-truth label. Where $y_n = 1$ when the image is drawn from the ground-truth label, and $y_n = 0$ when the image is generated from the segmentation network.

We propose to train the segmentation network via minimizing a multi-task loss function:

$$L_{seg} = L_{ce} + \lambda L_{adv} \quad (2)$$

where $L_{ce}$, $L_{adv}$ denote the multi-class cross entropy loss, the adversarial loss, respectively. λ represents a hyper-parameter for balancing the proportion of the $L_{ce}$ in multi-task loss function. $L_{ce}$ and $L_{adv}$ are respectively obtained by:

$$L_{ce} = -\sum_{h,w}\sum_{c \in C} Y_n^{(h,w,c)} \log\left(S(X_n)^{(h,w,c)}\right) \quad (3)$$

$$L_{adv} = -\sum_{h,w} \log\left(D(S(X_n))^{(h,w)}\right) \quad (4)$$

where $c$ is the number of categories in the dataset. With this adversarial loss, we first try to train the segmentation network to cheat the discriminator by maximizing the probability of the segmentation prediction being considered as the ground truth distribution.

## 4 Experiments

### 4.1 Dataset

We now detail the architectures we used for our preliminary experiment on the PASCAL VOC2012 [26] segmentation benchmark, which is a commonly used evaluation benchmark for semantic segmentation. It contains 20 objects except the background with annotations on daily captured images. As is common practice, we use the extra annotation set in SBD [27] for training, which provide a total of 10582 training images. We evaluate our models on the standard validation set with 1449 images.

### 4.2 Training settings

During the constant training process, we adopt the random scaling and cropping with size $319 \times 319$ for each image. The weights of the networks are initialized from the ResNet-101 model pre-trained on MS COCO dataset. In particular, we opt the Stochastic Gradient Descent (SGD) with Nesterov acceleration for the optimizer, where the momentum is set as 0.9 and the weight decay with factor $5 \times 10^{-4}$. The initial learning rate is set to $2.5 \times 10^{-4}$ and is decreased with polynomial decay with power of 0.9 as we reference the research of Chen et al. [9]. Besides, towards the goal of training the discriminator, we use an Adam solver with a learning rate of $10^{-4}$ and the same polynomial decay as the segmentation network. The momentum is set to 0.9 and 0.999. With each update of the segmentation network S, the generative network G will also be updated once time. We trained each model (in 50000 iterations with a batch-size of 11) on an Nvidia GeFore GTX1080Ti GPU using Pytorch [28] repository.

To show the capabilities of Seg-GANs we evaluate our method with several state-of-the-art algorithms, which including Luc et al., SegNet, FCN, DeepLab-v2 and DeepLab-v3. We use the PASCAL VOC2012 dataset as a basis, but augment the ground-truth with the goal to simulate prediction errors.

The MIoU is the most commonly used evaluation standard for semantic segmentation. It calculates the ratio of the intersection and union of the two sets, and finally averages the result. For semantic segmentation, the ratio between the predicted value and the true value is obtained. First, IoU is calculated on each class, and finally MIoU is obtained. To summarize, the MIoU can be defined as:

$$MIoU = \frac{1}{t+1}\sum_{i=0}^{t}\frac{p_{ii}}{\sum_{j=0}^{t}p_{ij}+\sum_{j=0}^{t}p_{ji}-p_{ii}} \tag{5}$$

where t+1 is the category number. $p_{ii}$, $p_{ij}$ and $p_{ji}$ denote true positive, false positive and false negative, respectively.

To obtain the best performance of the model, we set multiple hyper-parameter λ (see Equation (2)), which are 0.01, 0.02, 0.05, and 0.005, respectively. We trained each hyper-parameter training process in 50000 iterations, and each training process takes about 12 hours. The loss function curve under λ = 0.01 are revealed in Fig. 3.

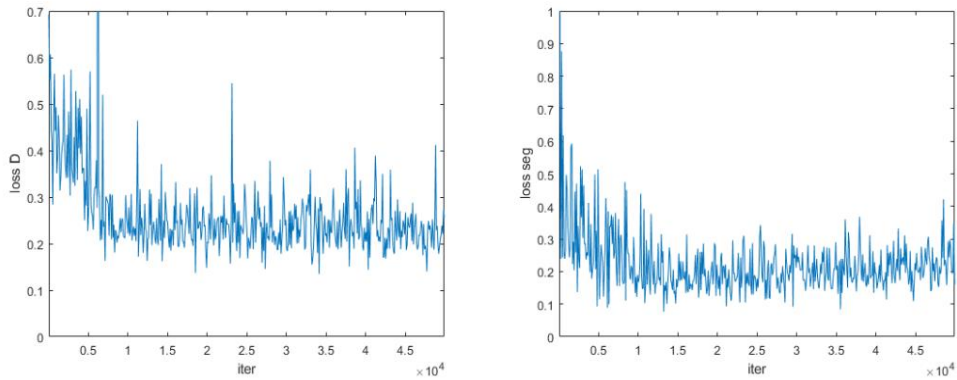

Figure 3. The loss function curve under $\lambda = 0.01$ of the segmentation network and discriminative network respectively.

Due to the use of a pre-trained model, the segmentation network converges after approximately 20,000 iterations, and the loss floats between 0.1 and 0.3. At this time, the loss value of the discriminator network fluctuates around 0.2. Towards the goal of obtaining the optimal model in this paper, we tested the MIoU curve between the models obtained from 10,000 iterations to the models obtained from 50,000 iterations on a test dataset with a step size of 1000, as shown in Fig. 4.

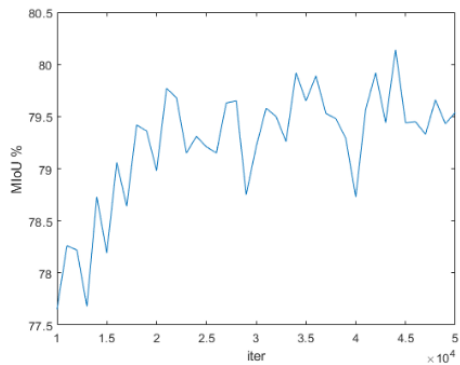

Figure 4. The curve of MIoU, from 10,000 iterations to 50,000 iterations, with a step size of 1000.

**4.2 Intuitive visual comparison**

To show the capabilities of Seg-GANs, in this section we evaluate our method with the state-of-the-art algorithm DeepLab-v2 [10]. In the generative network, we opt to DeepLab-v2 without CRFs post-processing as our basis segmentation network.

Figure 5 illustrate the comparative results of our method and the state-of-the-art algorithm DeepLab-v2. As shown in Fig. 5, we provide a visual comparison that our method is more detailed and completed than the DeepLab-v2 basic model. When segmenting some targets with more complex shapes, we can maintain the details without the big splits like DeepLab-v2. And the algorithm proposed in this paper abandoned to use the post-processing method, which is more efficient than DeepLab-v2.

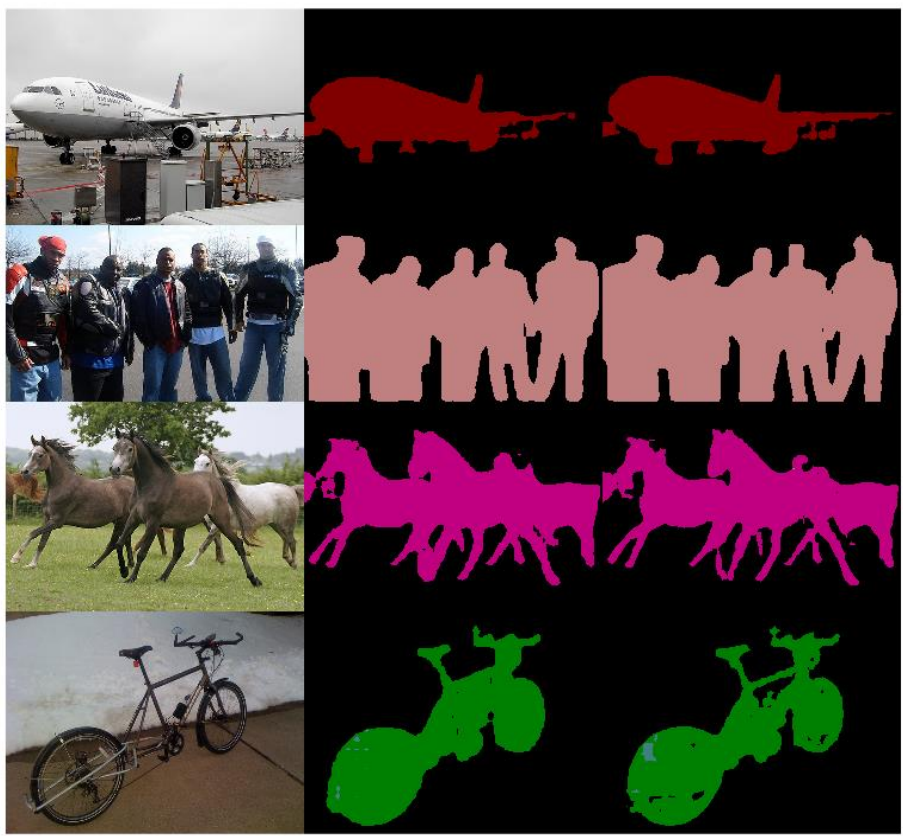

(a) Input   (b) DeepLab-v2   (C) Seg-GANs

Figure 5. The results of the DeepLab-v2 and Seg-GANs

## 4.3 Quantitative comparisons

For the sake of proving the efficiency of our proposed method, we compare Seg-GAN with several methods: SegNet, FCN, Luc et al., DeepLab-v2, DeepLab-v3. Using the exactly same dataset, we directly reference the results of SegNet, FCN in the original paper. We trained and tested DeepLab-v2 and DeepLab-v3 according to their paper. As shown in Table 1, our method always yields the highest scores. The results show that the results of proposed Seg-GAN outperform the other algorithms significantly.

Table 1 The average results of MIoU(%) on the VOC2012 dataset [26]

| Number | Methods | MIoU (%) |
|---|---|---|
| 1 | SegNet[1] | 60.50 |
| 2 | FCN[7] | 67.20 |
| 3 | Luc et al.[8] | 72.0 |
| 4 | DeepLab-v2 (without post-processing)[10] | 75.94 |
| 5 | DeepLab-v2 (with post-processing)[10] | 78.27 |
| 6 | DeepLab-v3[11] | 77.21 |
| 7 | Ours ($\lambda = 0.01$) | **80.14** |

The performance of the proposed model is affected by the hyper-parameter λ, which is a very sensitive parameter, and its value largely affects the accuracy of the segmentation. So we set different values, including 0.01, 0.02, 0.05, and 0.005. Table 2 gives the effect of different values on the performance of the proposed model.

Table 2 Hyper-parametric analysis

| $\lambda$ | MIoU (%) |
|---|---|
| 0.01 | **80.14** |
| 0.02 | 79.78 |
| 0.05 | 78.34 |
| 0.005 | 78.17 |

## 5 Conclusion

In this paper, a new end-to-end semantic segmentation model called Seg-GANs is proposed. In the Seg-GANs algorithm, the cascaded ConvCRFs is combined with GAN in the discriminative network adding the strong and valid assumption of conditional independence, and the cross-entropy error loss and adversarial loss are utilized to guide the training process through back propagation. Our generative network takes a remarkable basement segmentation network into consideration by integrating the existing segmentation network to realize the estimation of the original image. The discriminative network differentiates the segmentation results from the ground truth distribution and improves the details of the output images. The results show that the proposed Seg-GANs considerably improve the accuracy of the segmentation results. Our future work will mainly focus on exploring the potential of ConvCRFs in other structured applications such as instance segmentation and further improving the accuracy of small samples and reducing training time.

**Declarations:**

**1. Availability of data and material:** The dataset used during the current study is VOC2012 dataset [26], are available online or from the corresponding author on reasonable request.

**2. Competing interests:** The authors declare that they have no competing interests

**3. Fundings:** This work was supported in part by the National Natural Science Foundation of China (Grant No. 61403281), the Natural Science Foundation of Shandong Province (ZR2014FM002), China Postdoctoral Science Special Foundation Funded Project (2015T80717).

**4. Authors' contributions:** ZZ was a major contributor in writing the manuscript. And he analyzed and interpreted the entire framework of GANS with the help of ZW and Q.S. and H.Y. performed the coding and experiments. All authors read and approved the final manuscript.

**5. Acknowledgments:** Not applicable